\documentclass{xdenovo-report}
\reporttitle{DrugCoding: A Scientific Programming Agent \\ for Drug Discovery}
\reportshorttitle{DrugCoding | Technical Report}
\reportauthors{Junwei Hu\textsuperscript{\dag}, Zhuohao Lin\textsuperscript{\dag}, Kun Li\textsuperscript{\large\Letter}, and Hengnan Zhang\textsuperscript{\large\Letter}
\\ XDeNovo Labs, Shanghai, China}
\reportdate{29 October 2026}
\reportversion{1.0}

\usepackage{booktabs}
\usepackage{listings}
\usepackage{xcolor}
\usepackage{hyperref}
\usepackage{marvosym}

\usepackage{booktabs}
\usepackage{tabularx}
\usepackage[table]{xcolor}
\usepackage{array}

\definecolor{DCNavy}{HTML}{193B57}
\definecolor{DCAccent}{HTML}{167D91}
\definecolor{DCHeader}{HTML}{EDF3F7}
\definecolor{DCColumn}{HTML}{EAF4F6}
\definecolor{DCAverage}{HTML}{F2F5F7}

\newcolumntype{L}{>{\raggedright\arraybackslash}X}
\newcolumntype{C}{>{\centering\arraybackslash}X}

\newcolumntype{D}{%
  >{\columncolor{DCColumn}\centering\arraybackslash}X%
}

\definecolor{codegreen}{rgb}{0,0.6,0}

\lstdefinestyle{codestyle}{
    backgroundcolor=\color{gray!10},   
    basicstyle=\ttfamily\footnotesize,
    breaklines=true,
    numbers=left
}

\begin{document}
\makereporttitle

\begingroup
  \renewcommand{\thefootnote}{\dag}
  \footnotetext[0]{Junwei Hu and Zhuohao Lin contributed equally to this work and are interns at XDeNovo Labs.}
\endgroup

\begingroup
  \renewcommand{\thefootnote}{\Letter}
  \footnotetext[0]{Correspondence to Kun Li
  (\texttt{likun9876543210@163.com}) and Hengnan Zhang
  (\texttt{hengnanzh@xdenovoai.com}).}
\endgroup

\begin{abstract}
Drug discovery involves multiple interconnected research stages, including molecular design, property prediction, and virtual screening. These processes are characterized by complex workflows, interdependent computational tasks, and diverse specialized tools, placing substantial demands on the coordination of scientific models and computational resources. However, existing large language model (LLM)-based scientific agents still face challenges in autonomous planning of complex tasks, adaptive selection of specialized models, and reliable execution of multi-stage scientific computations, limiting their effectiveness in automating complex drug discovery workflows. To address these challenges, we propose \textbf{DrugCoding}, a scientific programming agent system that integrates scientific task planning, adaptive model routing, multi-agent collaboration, and programmatic tool execution. DrugCoding enables autonomous task planning and dynamic model adaptation for complex research objectives. It can flexibly invoke specialized scientific models, autonomously generate and execute code according to task requirements, and leverage multi-agent collaboration to support efficient execution and reliable validation of complex scientific tasks. To evaluate DrugCoding's performance in scientific tool invocation and multi-stage task execution, we conduct multi-turn tool-calling experiments across diverse language models and scientific tool interfaces. Experimental results demonstrate that DrugCoding achieves an average session-level Pass@1 of \textbf{97.35\%} on the full benchmark, outperforming the strongest baseline, Codex, by \textbf{9.24\%}. On the challenging subset, DrugCoding achieves \textbf{97.92\%}, exceeding the strongest baseline.
\end{abstract}

\begin{figure*}[t!]
    \centering
    \includegraphics[width=\textwidth]{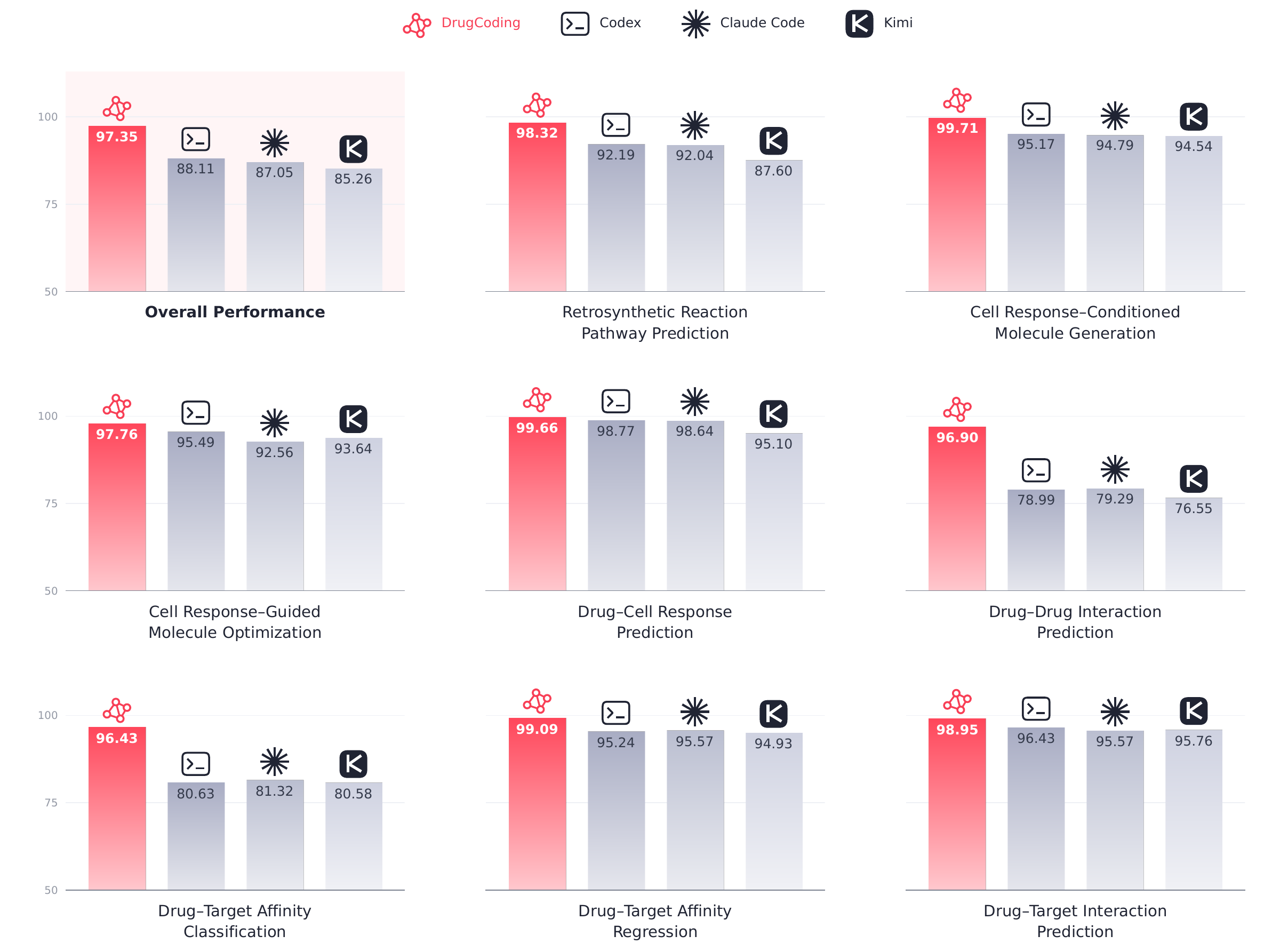}
    \caption{DrugCoding achieves superior performance across diverse drug discovery tasks. Pass@1(\%) comparison of DrugCoding and baseline agents. All results are averaged across 7 large language models: DeepSeek-V4-Flash, Claude Sonnet 5, Qwen 3.5, GPT-5.6-Sol, Qwen 3.8 Max, Qwen 3.5-35B-A3B, and Kimi K3.}
    \label{fig:task_accp}
\end{figure*}

\section*{Introduction}
\label{sec:introduction}

Drug discovery is a complex, multi-stage scientific process that increasingly relies on the coordinated use of computational models, specialized software, and biological databases. Research tasks such as target identification, molecular generation, property prediction, and structure-based screening involve distinct computational methods and often depend on the outputs of preceding stages~\cite{vamathevan2019applications}. Despite substantial advances in specialized scientific models, integrating these capabilities into coherent research workflows remains challenging. Different tools operate on heterogeneous data representations, such as protein sequences, molecular SMILES, and three-dimensional structures, and impose different requirements on computational environments, input parameters, and output formats~\cite{mendez2019chembl,corso2023diffdock}. Consequently, researchers must frequently perform data preprocessing, format conversion, model configuration, and result integration across multiple computational stages. More importantly, reliable execution requires not only connecting scientific tools but also managing task dependencies, selecting appropriate computational methods, and validating intermediate results. These challenges make it difficult to automate complex drug discovery workflows in a flexible, reproducible, and scalable manner.

Recent advances in large language models (LLMs) have created new opportunities for addressing these challenges through autonomous scientific agents. By combining natural language understanding, task planning, tool invocation, and code generation, LLM-based agents can coordinate specialized computational resources and execute scientific workflows with reduced manual intervention. Early systems such as ChemCrow demonstrated the potential of integrating LLM reasoning with domain-specific chemistry tools for molecular analysis, synthesis planning, and chemical research~\cite{bran2024augmenting}. Coscientist further combined autonomous planning, code execution, and experimental tools to conduct chemical research tasks~\cite{boiko2023autonomous}. These developments suggest that scientific agents can evolve beyond conversational assistance and isolated tool invocation toward more autonomous computational research.

Building on these advances, recent scientific agents have demonstrated increasingly sophisticated capabilities in drug discovery and biomedical research. DrugPilot introduces a parameterized reasoning architecture with structured memory and feedback mechanisms to support multi-stage drug discovery tasks and improve the reliability of scientific tool invocation~\cite{li2025drugpilot}. Biomni integrates retrieval-augmented planning, autonomous programming, and specialized biomedical tools to dynamically construct and execute workflows across diverse research tasks~\cite{huang2026biomni}. AgentD provides a modular framework that coordinates molecular generation, property prediction, structural analysis, and iterative molecular refinement for computational drug discovery~\cite{ock2026agentd}. More broadly, ToolUniverse provides a unified ecosystem for accessing and composing scientific tools~\cite{gao2025tooluniverse}, while Robin demonstrates multi-agent coordination between scientific hypothesis generation and experimental data analysis~\cite{ghareeb2026multiagent}. Collectively, these studies highlight the growing potential of LLM-based agents to integrate scientific reasoning with specialized computational capabilities.

Despite this progress, several challenges remain in developing flexible and reliable scientific programming agents for complex drug discovery workflows. First, scientific tasks often involve multiple interdependent computational stages whose execution order, input requirements, and intermediate results must be coordinated and updated as research progresses. Supporting such workflows requires explicit dependency management and adaptive task planning rather than relying solely on sequential tool invocation. Second, scientific problems can often be addressed by multiple specialized models with different applicability, predictive performance, and computational costs. Selecting and combining these models according to task requirements remains an important challenge beyond conventional tool identification. Third, scientific workflows frequently require computational operations that are not fully covered by predefined tools, creating a need to combine autonomous code generation with specialized model execution. Furthermore, integrating these capabilities with multi-agent collaboration, structured result management, and execution validation is essential for maintaining reliability across multi-stage scientific tasks. These challenges motivate a unified scientific programming framework capable of dynamically organizing computational resources and coordinating heterogeneous execution strategies.

To address these challenges, we propose DrugCoding, a scientific programming agent system for flexible and reliable computational drug discovery. DrugCoding integrates scientific task planning, adaptive model routing, autonomous programming, and multi-agent collaboration into a unified framework. It decomposes research objectives into dependency-aware tasks, selects scientific models based on applicability, output quality, and computational cost, and dynamically coordinates specialized tools and code execution. Multi-agent collaboration enables parallel exploration and result integration, while persistent memory and validation mechanisms support execution traceability, error detection, and failure recovery. We evaluate DrugCoding through multi-turn tool-calling experiments across diverse language models and scientific interfaces. DrugCoding achieves an average Pass@1 of 97.35\% on TCDD, a drug discovery benchmark introduced by DrugPilot\cite{li2025drugpilot}, outperforming the strongest evaluated baseline by 9.24\%. On the top 10\% most challenging tasks in TCDD, DrugCoding achieves a Pass@1 of 97.92\%, outperforming the strongest baseline by 1.31\%. These results demonstrate the effectiveness of DrugCoding in reliable scientific tool invocation across heterogeneous computational interfaces. The main contributions are summarized as follows:

\begin{enumerate}
    \item DrugCoding is proposed as a scientific programming agent framework for drug discovery, integrating autonomous task planning, specialized model invocation, code generation, and scientific workflow execution to support flexible and reliable execution of complex scientific tasks.

    \item An adaptive model routing mechanism and a multi-agent collaboration framework are developed to dynamically select specialized scientific models according to task requirements, model applicability, and computational cost, enabling task decomposition, parallel exploration, and coordinated multi-stage execution.

    \item Experimental results demonstrate that DrugCoding achieves an average session-level Pass@1 of 97.35\% on the full benchmark and 97.92\% on the challenging subset, outperforming the strongest baselines by 9.24\% and 1.31\%, respectively, while maintaining consistently strong performance across eight tool-specific evaluations.
\end{enumerate}

\textbf{Results.} DrugCoding achieves an average Pass@1 of 97.35\% across 7 language-model configurations, compared with 88.11\% for Codex, 87.05\% for Claude Code, and 85.26\% for Kimi, as shown in Figure~\ref{fig:task_accp}. DrugCoding also obtains the highest average scores across all eight scientific subtasks, with particularly large advantages in drug-drug interaction prediction and drug-target affinity classification. We refer the reader to \emph{Evaluation} Section for the detailed comparisons, ablation experiments, failure analyses, and case studies.

\section*{DrugCoding System Architecture}
\label{sec:drugcoding_architecture}

Computational drug discovery connects molecular design, property prediction, and structural analysis through workflows with heterogeneous inputs and interdependent results. Coordinating these operations requires both scientific decisions about what to compute and explicit procedures for carrying those decisions into execution. DrugCoding is a scientific programming agent framework that translates natural-language research objectives and supplied data into structured, traceable computational workflows. Its architecture combines two central features. \textbf{{Multi-agent collaboration}} distributes context resolver, workflow planning, orchestration, model routing, scientific execution, and evidence synthesis among role-specific components. \textbf{{Adaptive model routing}} connects each scientific task to an appropriate model and its execution procedure. The following sections first describe the agents, then explain how model selection is translated into verified computation and scoped continuation.


\begin{figure}[t!]
    \centering
    \includegraphics[width=\textwidth]{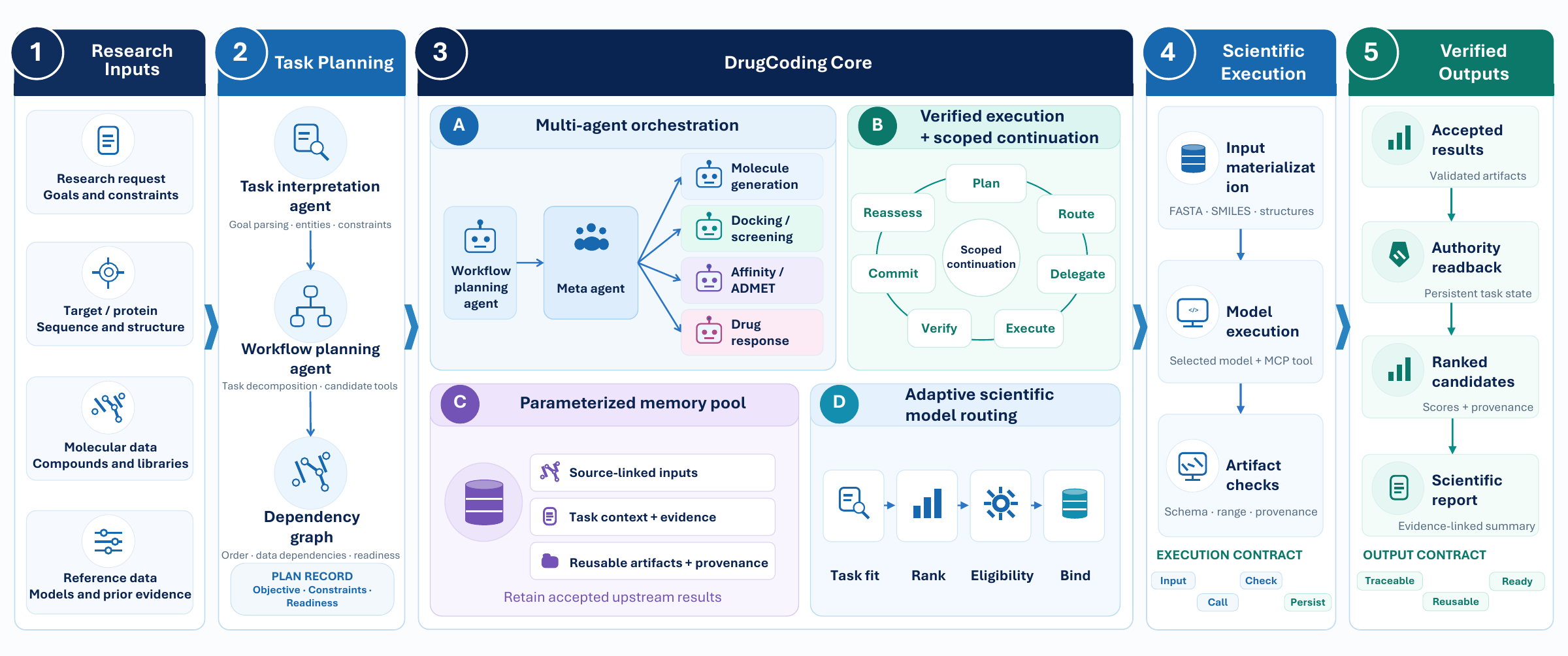}
    \caption{DrugCoding system architecture across five stages. (1) Research inputs collect the research request, target and protein information, molecular data, and reference evidence. (2) Task planning interprets the objective and constructs a dependency graph. (3) The DrugCoding core coordinates the workflow. (4) Scientific execution materializes inputs, invokes selected models, and checks returned artifacts. (5) Verified outputs persist accepted results for readback, ranking, and scientific reporting.}
    \label{fig:drugcoding_architecture}
\end{figure}

\subsection*{Multi-agent collaboration}
\label{sec:drugcoding_architecture_3}

DrugCoding assigns distinct reasoning responsibilities to agents while retaining explicit control over scientific inputs, execution state, and accepted results. A logical task specifies a scientific operation and its dependencies. An execution binding specifies the selected model, input resources, and execution procedure. An accepted result associates the resulting artifacts with their provenance. The host manages input preparation, plan acceptance, readiness assessment, and execution interfaces; authority stores accepted plans, bindings, and results. These programmatic components provide the shared state through which the agents exchange work.

\paragraph{Meta agent.}
The meta agent coordinates the task lifecycle from plan integration to result collection. It receives the research request, structured inputs, and the workflow planning agent's proposal through an entry point that performs programmatic consistency checks. Planning integration replays this proposal through the shared lifecycle without issuing a second planning language-model request, and the host accepts the resulting plan. During execution, the meta agent delegates each ready stage to the corresponding scientific execution agent and receives the stage-level result projection. The same component therefore connects planning and execution through a shared lifecycle. Its available tools change with the stage: planning integration exposes no scientific execution tools, whereas execution exposes the relevant delegation interface.

\paragraph{Context resolver agent.}
The context resolver agent establishes the scientific entities and constraints used by subsequent agents. It runs in a dedicated session through a separate model call before the workflow planning agent is invoked. Its context contains the current request, supplied materials, and an eligible history projection restricted to user messages. It extracts molecules, proteins, cellular contexts, requested outputs, and their relationships without invoking scientific tools. Programmatic checks validate the returned structure and associate the extracted entities with their source records. When identical protein-sequence records occur in history, the host chooses the newest record by its authenticated creation or source timestamp, records the selected record identifier and provenance, and retains older records as non-selected evidence. The resulting referenceable input objects preserve scientific identities and pass to planning and coordination.

\paragraph{Workflow planning agent.}
The workflow planning agent receives the research objective and validated input projection in its own session, distinct from the context resolver agent's session. Explicitly requested memory evidence can also enter planning, with its attribution retained. The agent produces a structured proposal specifying scientific capabilities, input references, predecessor relationships, and conditions for expanding downstream stages. Each atomic task describes an operation and the inputs or upstream outputs it requires. The proposal also separates requested scientific outputs from computational preparation. For example, docking from a protein sequence may require receptor and pocket preparation without adding complex-structure prediction as a separate research objective. The meta agent integrates the proposal, and the host resolves its references, assesses readiness, and submits the accepted plan to authority.

\paragraph{Model routing agent.}
The model routing agent identifies suitable scientific models for each task with competing candidates. It receives the task context, candidate descriptions, and supplied scientific evidence, then returns a ranked recommendation with supporting rationale. This recommendation informs the execution layer, which determines candidate eligibility and establishes the final binding. The agent's responsibility is scientific comparison; the transition from recommendation to executable model is described in the adaptive routing section.

\paragraph{Scientific execution agent.}
The scientific execution agent carries out the scientific invocation assigned by the meta agent. It receives the current stage's accepted inputs, constraints, and observations through dedicated context providers, together with the permitted scientific tool interface. It then invokes that interface and returns an execution record. The resulting output passes through verification and persistence before its accepted projection returns to the meta agent. This division keeps generated explanations distinct from records used for downstream computation.

\paragraph{Evidence synthesis agent.}
The evidence synthesis agent generates a Markdown report from accepted scientific records, field descriptions, and artifact references. Each authoring attempt uses a dedicated session with no scientific tools, leaving accepted computations unchanged. An empty response, incomplete generation, or invocation error permits at most one fresh authoring attempt. This attempt uses the same scientific evidence with an instruction to return a complete, concise report. Available partial text is retained with its status; failure to obtain report text is reported explicitly. The current report path ends with authoring and status recording, without an independent language-model review.

\subsection*{Adaptive model routing}
\label{sec:drugcoding_architecture_4}

A scientific capability can be implemented by models with different applicability, input requirements, and output semantics. DrugCoding therefore separates the task's scientific objective from the model and execution procedure used to realize it. Adaptive routing links the model routing agent's recommendation to execution eligibility, a final model binding, and invocation by the scientific execution agent. In this section, SMILES denotes the Simplified Molecular Input Line Entry System, ADMET denotes absorption, distribution, metabolism, excretion, and toxicity, AUC denotes the area under the dose--response curve, SDF denotes a Structure Data File, and MCP denotes the Model Context Protocol. Table~\ref{tab:scientific_capabilities_models} summarizes the supported interfaces, their inputs and outputs, and the models registered for each capability. Routing concerns these scientific models rather than the general-purpose language models used by the agents.

\begin{table*}[t]
\centering
\caption{\textbf{Scientific capabilities and registered models in DrugCoding.}
Inputs and outputs are specified at the scientific-object level, while model-specific preprocessing converts them into native execution formats when required. Small-molecule inputs are represented by SMILES unless a 3D structure is explicitly required.}
\label{tab:scientific_capabilities_models}

\small

\begin{tabularx}{\linewidth}{
>{\raggedright\arraybackslash}p{0.205\linewidth}
>{\raggedright\arraybackslash}X
>{\raggedright\arraybackslash}X
>{\raggedright\arraybackslash}p{0.175\linewidth}
}

\toprule

\rowcolor{DCHeader}
\textcolor{DCNavy}{\bfseries Capability} &
\textcolor{DCNavy}{\bfseries Input} &
\textcolor{DCNavy}{\bfseries Output} &
\textcolor{DCNavy}{\bfseries Registered models} \\

\midrule

\textcolor{DCNavy}{\bfseries Drug-target affinity prediction} &
Molecule (SMILES; e.g., \texttt{CC(=O)OC1=CC=CC=C1C(=O)O}); target protein sequence; optional protein features &
Predicted drug--target binding affinity, reported as $pK_d$ &
\textcolor{DCAccent}{
DMFF-DTA~\cite{he2025dual};
DTIAM~\cite{lu2025dtiam};
TSEDTA~\cite{sun2026tsedta}
} \\

\rowcolor{DCHeader!30}
\textcolor{DCNavy}{\bfseries ADMET prediction} &
Molecule (SMILES) &
Predicted absorption, distribution, metabolism, excretion, and toxicity endpoints &
\textcolor{DCAccent}{
ADMET-AI~\cite{swanson2024admet}
} \\

\textcolor{DCNavy}{\bfseries Structure-based virtual screening} &
Target protein pocket (3D structure); molecular library (SMILES or SDF) &
Ranked candidate molecules with model-specific screening scores &
\textcolor{DCAccent}{
DrugCLIP~\cite{gao2023drugclip}
} \\

\rowcolor{DCHeader!30}
\textcolor{DCNavy}{\bfseries Protein-ligand complex prediction} &
Target protein sequence; ligand molecule (SMILES) &
Predicted protein--ligand complex structure, with confidence and affinity estimates when available &
\textcolor{DCAccent}{
Boltz-2~\cite{passaro2025boltz}
} \\

\textcolor{DCNavy}{\bfseries Protein-ligand docking} &
Ligand molecule (SMILES or 3D structure); receptor or binding-pocket structure &
Ranked ligand binding poses with confidence or docking scores &
\textcolor{DCAccent}{
DiffDock~\cite{corso2022diffdock};
KarmaDock~\cite{zhang2023efficient}
} \\

\rowcolor{DCHeader!30}
\textcolor{DCNavy}{\bfseries Pocket-conditioned molecular generation} &
Target protein pocket (3D structure); optional reference ligand or molecular core &
Generated molecules represented as canonical SMILES and/or 3D structures &
\textcolor{DCAccent}{
MolCRAFT~\cite{qu2024molcraft};
ED2Mol~\cite{li2025electron};
Pocket2Mol~\cite{peng2022pocket2mol}
} \\

\textcolor{DCNavy}{\bfseries Cell-line drug-response prediction} &
Molecule (SMILES); target cell line; matched molecular profile (e.g., gene expression) &
Predicted drug response, reported as normalized dose--response AUC, with ensemble uncertainty &
\textcolor{DCAccent}{
PharmaFormer~\cite{zhou2025pharmaformer}
} \\

\bottomrule

\end{tabularx}
\begin{minipage}{\linewidth}
\footnotesize
\emph{Note:} The registered models represent valuable contributions to computational drug discovery. Their original authors and developers are gratefully acknowledged. These models are used in this study solely for research evaluation, with no commercial use.
\end{minipage}
\end{table*}

\paragraph{Scientific recommendation.}
For tasks with competing candidates, the model routing agent compares the catalogued models in an independent session. The supplied context describes the task, candidate applicability, and available performance evidence. The agent prioritizes primary performance evidence, using task fit when that evidence is unavailable, incomparable, or tied. Its recommendation scores express relative preference rather than calibrated probabilities of success. Comparison uses the supplied evidence without real-time external retrieval. Deployment convenience, caching, latency, and resource cost do not enter this scientific ranking. Capabilities with a fixed single candidate proceed without competitive ranking.

\paragraph{Execution eligibility and binding.}
The execution layer translates the recommendation into a model that can run with the task's available inputs and resources. It checks input requirements, data contracts, permitted executors, adapter compatibility, and runtime availability. An eligible manual choice takes precedence; an ineligible choice is rejected. Automatic selection chooses the highest-ranked eligible candidate, which may differ from the first scientific recommendation. If no recommended candidate is eligible, no execution binding is created. When scientific advice is absent, eligible candidates follow the catalog's deterministic order. Tasks awaiting required context remain pending until comparison and binding can proceed.

The final binding records the selected model, checkpoint resources, input builder, and executor. These components determine how the scientific task is translated into native inputs and runtime arguments. Preparation and dispatch consume this binding without selecting the model again inside the worker. The execution receipt is subsequently checked against the binding, preserving the distinction between a recommended model and the model actually executed.

\paragraph{Scientific execution agent and tool invocation.}
The scientific execution agent connects the accepted binding to a scientific tool exposed through the model context protocol~\cite{protocol2026model} (MCP). The meta agent provides the delegation instruction, while context providers supply the current stage's inputs, constraints, and observations. The scientific execution agent creates a separate session object that shares task state with its parent. Context delivery determines which information is presented to the model, so shared state does not imply unrestricted access to the full conversation. The exposed tool schema defines the permitted invocation, and surrounding programs enforce the stage's input scope and invocation limits.

Input preparation precedes the scientific call and follows the selected model's requirements. The host resolves input references, builders construct native artifacts, and adapters translate the binding into worker arguments and runtime configuration. Each worker uses its designated interpreter, dependencies, and model resources, separately from the agent and MCP control environment. Scientific coding in this workflow comprises the programmatic composition of these transformations and execution steps. This arrangement allows the agent to request computation without regenerating molecular representations, file paths, checkpoint locations, or device settings in its tool arguments.

\paragraph{Verified execution and scoped continuation.}
The result-verification layer checks each scientific output against its task contract before downstream use. It validates structure, entity correspondence, numerical fields, artifact availability, and consistency with the selected model. These checks are deterministic and do not create an additional agent session. A failed post-tool check produces an explicit error rather than an accepted scientific result. The middleware records the verification failure and raises an execution failure. Recognized stage-local failures are recorded against the affected stage and block its dependent tasks; failures outside that category terminate the affected run. The verification layer does not generate corrective code, select replacement models, or initiate a new scientific invocation.

A failed stage emits a structured signal containing the affected stage, violated contract, and execution state. The meta agent determines whether the failure is local, retryable, or blocking. When a bounded correction is available, only the affected stage is redelegated while accepted upstream results remain unchanged. If required evidence is unavailable or execution state is unknown, the system reports an unresolved state rather than claiming completion.

Accepted upstream outputs support continuation within the existing plan as downstream requirements become satisfied. The host constructs the affected inputs and reassesses readiness together with any outstanding user choices. For example, generated molecules can populate drug-response tasks once the required cell context is available. Updating these inputs preserves the original research scope and completed computations. Stages awaiting prerequisites remain pending, while independent later stages can undergo readiness assessment. The execution workflow processes scientific stages in an ordered delegation--execution--verification chain.

\paragraph{Capability-level execution interface.}
For each scientific capability, the execution layer prepares model-specific inputs, invokes the selected implementation, validates the returned contract, and stores the accepted result for downstream stages. The same interface applies across affinity prediction, virtual screening, docking, molecular generation, and cellular-response prediction.

\paragraph{Result return and downstream use.}
After the scientific execution agent finishes, task-specific checks and normalization produce the typed scientific result. Authority stores the result, and readback supplies the accepted output for downstream tasks, reporting, and the meta agent's stage-level result projection. Connection establishment, tool invocation, worker completion, and result acceptance remain distinct events. The result record associates artifacts with the task and executed model while preserving their scientific meaning. Virtual-screening scores, docking scores, binding-affinity predictions, and cellular-response predictions consequently retain their native semantics. Together, recommendation, binding, execution, and verified return connect a common scientific task interface to model-specific computation.

\section*{Evaluation}

\subsection*{Experimental Settings}
\paragraph{Evaluation data.}
The Tool-Calling Dataset for Drug Discovery (TCDD), introduced by DrugPilot~\cite{li2025drugpilot}, is used for evaluation. TCDD distinguishes three tool-calling scenarios: simple function, where only one function is provided; multiple function, where the agent selects one function from eight available tools; and multi-turn function, where successive user requests may require different tool calls and the reuse or revision of previously supplied arguments. We focus on the more demanding multi-turn scenario to evaluate DrugCoding's ability to maintain context, select tools, and preserve scientific arguments across turns. Accordingly, the full evaluation benchmark comprises the complete multi-turn portion of TCDD, including all 1,250 sessions in the released data, corresponding to 2,500 action-level inputs across eight tool interfaces, termed \textbf{M-TCDD}. 
To reduce the computational cost of comparisons with Claude Science and Rosalind Workbench, we additionally construct a challenging subset containing approximately 10\% of these sessions and evaluate DrugCoding on the same subset, termed \textbf{M-10\%-TCDD}. We group complete sessions by the reference tool used in their final action and rank sessions within each group using a model-independent difficulty score. This score is a weighted average of within-group feature percentiles, incorporating contextual references, tool switching, argument length and structure, and conversation and schema complexity. We select the highest ranked sessions from each group, rounding the 10\% quota upward and retaining at least five sessions per group. This procedure produces 128 complete sessions with 256 action-level inputs while preserving their original conversational context.

\paragraph{Baselines and backbone models.}
DrugCoding is compared with three coding agent configurations accessed through their command-line interfaces (CLIs): Codex~\cite{openai_codex_cli}, Claude Code~\cite{anthropic_claude_code}, and Kimi~\cite{moonshot_kimi_cli}. Each agent is configured with the designated model provider and adapted to the benchmark through protocol conversion and structured-output parsing. The comparisons therefore concern these configured agent pipelines rather than the default commercial products. Seven backbone model configurations are evaluated on M-TCDD: DeepSeek-V4-Flash~\cite{deepseek_v4}, Claude Sonnet 5~\cite{anthropic_sonnet5}, Qwen 3.5~\cite{qwen35}, GPT-5.6-Sol~\cite{openai_gpt56sol}, Qwen 3.8 Max~\cite{qwen38}, Qwen 3.5-35B-A3B~\cite{qwen35_35b_a3b}, and Kimi K3~\cite{moonshot_kimi_k3}. Each evaluation input contains the supplied conversation history and tool schemas, and a structured prediction is returned for scoring under a 300-second timeout. 
Additional comparisons with two scientific research agents, Claude Science~\cite{anthropic_claude_science} and Rosalind Workbench~\cite{openai_rosalind_workbench}, are conducted on M-10\%-TCDD. Claude Science is accessed through its web interface, whereas Rosalind Workbench is evaluated through its skill-based configuration on Codex CLI. Both receive the benchmark conversation context and task specifications, and their structured submissions are assessed using the same scoring criteria. These experiments are organized under the GPT-5.6-Sol~\cite{openai_gpt56sol}, GPT-6-Sol~\cite{openai_gpt6sol}, and GPT-6.1-Sol~\cite{openai_gpt61sol} configuration labels.

\paragraph{Evaluation metrics and implementation.}
Agent task completion capability is evaluated using Pass@1, defined as the proportion of tasks successfully completed by a single final submission from the evaluated agent pipeline. Success is determined by whether the submission satisfies the reference-specified requirements, rather than by the agent's natural language claim of completion. In the present benchmark, the evaluated deliverable is a structured prediction containing the required operations and their complete normalized inputs. For the overall evaluation, each multi-turn session receives a binary score: it passes only if the submitted predictions match the reference at every required step, including operation identity, ordering, and all argument values. Missing or additional operations, incorrect ordering, incomplete outputs, and argument mismatches result in failure. For the subtask analysis in Figure~\ref{fig:task_accp}, the same exact-match criterion is applied to each individual action, and the resulting scores are averaged across seven language model configurations.

\begin{table*}[t]
\centering
\caption{\textbf{Performance comparison on the M-TCDD benchmark.} Pass@1 (\%) of DrugCoding and three general-purpose coding agents across seven backbone LLMs. The best and second-best distinct scores in each row are shown in bold and underlined,
respectively.}
\label{tab:full_multiturn}

\begingroup
\small
\renewcommand{\arraystretch}{1.18}
\setlength{\tabcolsep}{6pt}

\begin{tabularx}{\linewidth}{
>{\hsize=1.45\hsize}L
>{\hsize=0.90\hsize}D
>{\hsize=0.82\hsize}C
>{\hsize=1.03\hsize}C
>{\hsize=0.80\hsize}C
}

\toprule

\rowcolor{DCHeader}
\textcolor{DCNavy}{\bfseries Backbone LLM} &
\cellcolor{DCHeader}\textcolor{DCAccent}{\bfseries DrugCoding} &
\textcolor{DCNavy}{\bfseries Codex} &
\textcolor{DCNavy}{\bfseries Claude Code} &
\textcolor{DCNavy}{\bfseries Kimi} \\

\midrule

DeepSeek-V4-Flash &
\textcolor{DCNavy}{\textbf{97.36}} &
81.28 &
80.00 &
\underline{87.28} \\

Claude Sonnet 5 &
\textcolor{DCNavy}{\textbf{96.64}} &
90.16 &
87.92 &
\underline{90.80} \\

Qwen 3.5 &
\textcolor{DCNavy}{\textbf{95.36}} &
\underline{84.56} &
83.68 &
74.00 \\

GPT-5.6-Sol &
\textcolor{DCNavy}{\textbf{98.40}} &
83.52 &
\underline{84.00} &
81.20 \\

Qwen 3.8 Max &
\textcolor{DCNavy}{\textbf{96.32}} &
\underline{94.00} &
92.72 &
92.48 \\

Qwen 3.5-35B-A3B &
\textcolor{DCNavy}{\textbf{97.76}} &
\underline{85.92} &
82.72 &
74.24 \\

Kimi K3 &
\textcolor{DCNavy}{\textbf{99.60}} &
97.36 &
\underline{98.32} &
96.80 \\

\midrule

\rowcolor{DCAverage}
\textcolor{DCNavy}{\bfseries Average} &
\cellcolor{DCColumn}\textcolor{DCNavy}{\bfseries 97.35} &
\underline{88.11} &
87.05 &
85.26 \\

\bottomrule

\end{tabularx}
\endgroup
\end{table*}


\begin{table*}[t]
\centering
\caption{\textbf{Performance comparison on the top 10\% most challenging TCDD tasks (M-10\%-TCDD).} Pass@1 (\%) of DrugCoding and scientific research agents across three backbone LLMs. The best and second-best distinct scores in each row are shown in bold and underlined, respectively.}
\label{tab:hard_multiturn}

\begingroup
\small
\renewcommand{\arraystretch}{1.18}
\setlength{\tabcolsep}{6pt}

\begin{tabularx}{\linewidth}{
>{\hsize=1.05\hsize}L
>{\hsize=0.90\hsize}D
>{\hsize=1.00\hsize}C
>{\hsize=1.05\hsize}C
}

\toprule

\rowcolor{DCHeader}
\textcolor{DCNavy}{\bfseries Backbone LLM} &
\cellcolor{DCHeader}\textcolor{DCAccent}{\bfseries DrugCoding} &
\textcolor{DCNavy}{\bfseries Claude Science} &
\textcolor{DCNavy}{\bfseries Rosalind Workbench} \\

\midrule

GPT-5.6-Sol &
\textcolor{DCNavy}{\textbf{96.09}} &
\textbf{96.09} &
\underline{94.53} \\

GPT-6-Sol &
\textcolor{DCNavy}{\textbf{98.44}} &
\underline{97.66} &
95.31 \\

GPT-6.1-Sol &
\textcolor{DCNavy}{\textbf{99.22}} &
96.09 &
\underline{98.44} \\

\midrule

\rowcolor{DCAverage}
\textcolor{DCNavy}{\bfseries Average} &
\cellcolor{DCColumn}\textcolor{DCNavy}{\bfseries 97.92} &
\underline{96.61} &
96.09 \\

\bottomrule

\end{tabularx}
\endgroup
\end{table*}


\subsection*{Performance Across Models and Agent Baselines}
To assess DrugCoding's performance across language models and agent baselines, we evaluated it on the full and challenging multi-turn benchmarks. On the M-TCDD benchmark, DrugCoding achieved the highest Pass@1 across all seven model configurations, as shown in Table~\ref{tab:full_multiturn}. Its average score of 97.35\% exceeded Codex, Claude Code, and Kimi by absolute margins of 9.24\%, 10.30\%, and 12.09\%, respectively. Scores ranged from 95.36\% to 99.60\%, showing consistently high performance across the evaluated configurations.
The absolute advantage over the strongest baseline ranged from 1.28\% to 14.40\%. The largest gain occurred with GPT-5.6-Sol, where DrugCoding achieved 98.40\%, compared with 84.00\% for Claude Code. With Kimi K3, DrugCoding reached its highest score of 99.60\%, exceeding Claude Code by an absolute margin of 1.28\%. These comparisons show that the magnitude of the advantage depended on the model configuration; they do not isolate the contributions of individual architectural components.
On the M-10\%-TCDD benchmark, as shown in Table~\ref{tab:hard_multiturn}, DrugCoding achieved an average Pass@1 of 97.92\%, exceeding Claude Science and Rosalind Workbench by absolute margins of 1.31\% and 1.83\%, respectively. DrugCoding tied Claude Science at 96.09\% under the GPT-5.6-Sol configuration and led under GPT-6-Sol and GPT-6.1-Sol, scoring 98.44\% and 99.22\%, respectively. 
These results indicate that DrugCoding maintained high Pass@1 across the evaluated model configurations and achieved the highest average performance against both general-purpose coding agents and scientific research agents.

\begin{table}[ht]
\centering
\caption{Ablation study comparing single-agent and multi-agent DrugCoding using Pass@1 (\%). The better result for each model is bold. Average denotes the arithmetic mean across models.}
\label{tab:single_multi_agent}

\small
\renewcommand{\arraystretch}{1.18}
\setlength{\tabcolsep}{8pt}

\begin{tabular}{lrr}
\toprule

\rowcolor{DCHeader}
\textcolor{DCNavy}{\bfseries Model} &
\textcolor{DCNavy}{\bfseries Single-Agent} &
\textcolor{DCAccent}{\bfseries Multi-Agent} \\

\midrule

DeepSeek-V4-Flash &
95.76 &
\cellcolor{DCColumn}\textcolor{DCNavy}{\textbf{97.36}} \\

Claude Sonnet 5 &
\textbf{98.80} &
\cellcolor{DCColumn}96.64 \\

Qwen 3.5 &
\textbf{96.96} &
\cellcolor{DCColumn}95.36 \\

GPT-5.6-Sol &
92.16 &
\cellcolor{DCColumn}\textcolor{DCNavy}{\textbf{98.40}} \\

Qwen 3.8 Max &
\textbf{97.60} &
\cellcolor{DCColumn}96.32 \\

Qwen 3.5-35B-A3B &
96.88 &
\cellcolor{DCColumn}\textcolor{DCNavy}{\textbf{97.76}} \\

Kimi K3 &
99.36 &
\cellcolor{DCColumn}\textcolor{DCNavy}{\textbf{99.60}} \\

\midrule

\rowcolor{DCAverage}
\textcolor{DCNavy}{\bfseries Average} &
96.79 &
\cellcolor{DCColumn}\textcolor{DCNavy}{\textbf{97.35}} \\

\bottomrule
\end{tabular}
\end{table}


\subsection*{Effect of Multi-Agent Coordination}
To assess the effect of the multi-agent configuration, we compared single-agent and multi-agent DrugCoding across seven language-model configurations using Pass@1. Table~\ref{tab:single_multi_agent} demonstrates that the multi-agent version achieved an average score of 97.35\%, compared with 96.79\% for the single-agent version, an absolute improvement of 0.56\%.
The multi-agent version outperformed its single-agent counterpart in four of the seven configurations. The largest gain occurred with GPT-5.6-Sol, where Pass@1 increased from 92.16\% to 98.40\%, an absolute improvement of 6.24\%. Smaller gains were observed with DeepSeek-V4-Flash, Qwen 3.5-35B-A3B, and Kimi K3, ranging from 0.24\% to 1.60\%. Conversely, the single-agent version performed better with Claude Sonnet 5, Qwen 3.5, and Qwen 3.8 Max, with absolute advantages ranging from 1.28\% to 2.16\%.
Overall, the multi-agent configuration provided a modest average improvement, but its benefits varied across models. These results support evaluating the execution configuration jointly with the underlying language model rather than assuming a uniform benefit from additional agents. 

\begin{figure*}[ht]
    \centering
    \includegraphics[width=0.9\textwidth]{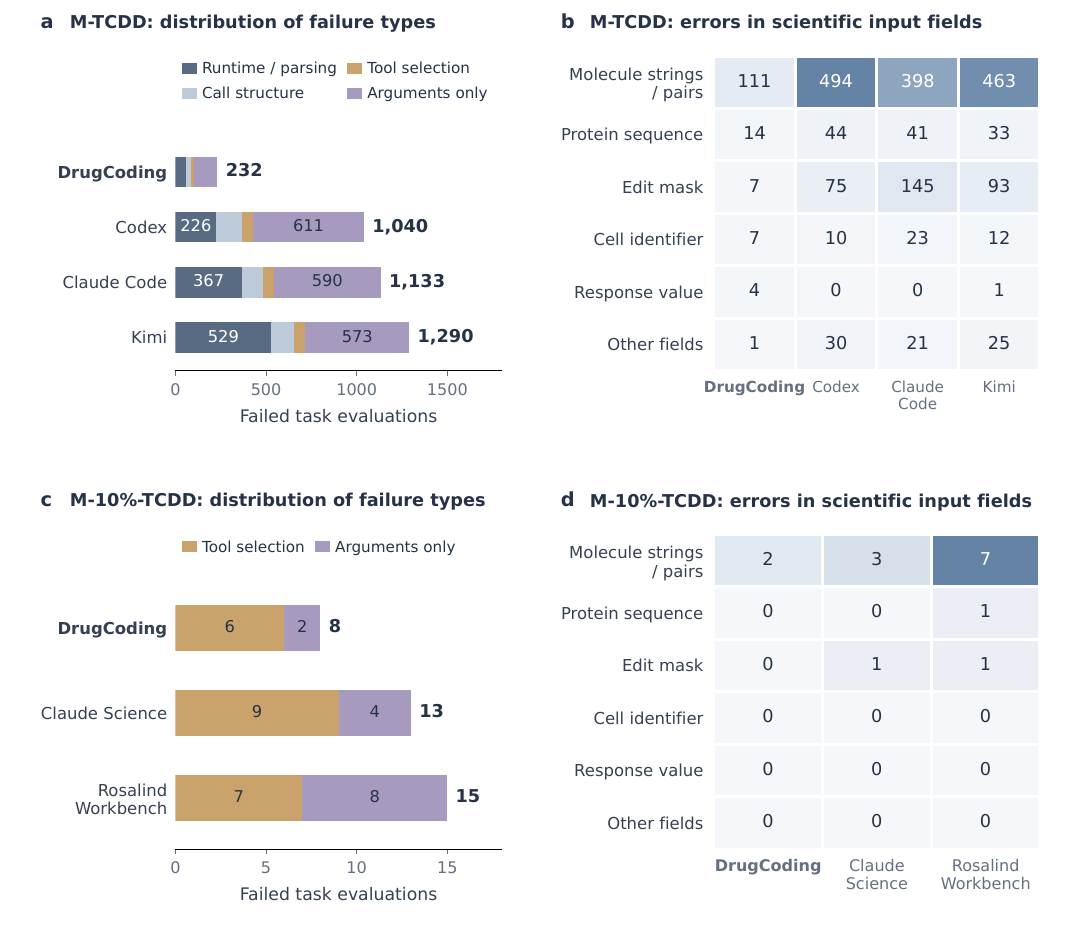}
    \caption{Failure patterns on the M-TCDD and M-10\%-TCDD benchmarks. \textbf{a,c}, Failed evaluations grouped into mutually exclusive categories using the priority order: runtime or parsing error, call-structure error, incorrect tool selection, and argument mismatch. \textbf{b,d}, Counts of actions with argument-field mismatches despite correct tool sequences; field groups may overlap.}
    \label{fig:combined_failure_analysis}
\end{figure*}

\subsection*{Failure Patterns Across Agent Configurations}
To examine where DrugCoding's performance advantage arose, archived predictions were analyzed on the M-TCDD and M-10\%-TCDD benchmarks in Figure~\ref{fig:combined_failure_analysis}. We grouped failed sessions into mutually exclusive categories and examined which scientific inputs differed from the reference when the predicted tool sequence was correct. This analysis distinguished failures in selecting and organizing operations from failures in preserving their required inputs.

On the M-TCDD benchmark, DrugCoding recorded 232 failed evaluations, compared with 1,040 for Codex, 1,133 for Claude Code, and 1,290 for Kimi, as shown in Figure~\ref{fig:combined_failure_analysis}a. The largest differences concerned argument mismatches and runtime or parsing errors. DrugCoding had 132 argument-only failures, compared with 611, 590, and 573 for the three baselines, respectively. It also recorded fewer runtime or parsing failures: 58, compared with 226, 367, and 529. The argument-field analysis localized much of the difference to molecular strings and editing masks in Figure~\ref{fig:combined_failure_analysis}b. Molecular-string or molecule-pair mismatches affected 111 DrugCoding actions, compared with 494, 398, and 463 baseline actions; editing-mask mismatches affected seven DrugCoding actions, compared with 75, 145, and 93. DrugCoding also recorded fewer tool-selection failures, with 14 compared with 60, 60, and 63 for the three baselines. These results associate its performance advantage with fewer operation-selection errors, more accurate preservation of scientific inputs, and more reliable structured outputs.

On the M-10\%-TCDD benchmark, the remaining failures concerned tool selection and argument correctness, with no recorded runtime, parsing, or call-structure failures, as shown in Figure~\ref{fig:combined_failure_analysis}c. DrugCoding failed eight evaluations, compared with 13 for Claude Science and 15 for Rosalind Workbench. Of these, only two DrugCoding failures involved arguments alone, compared with four and eight for the scientific agents. Its field-level mismatches were confined to molecular strings or pairs, whereas the baselines also exhibited editing-mask or protein-sequence errors in Figure~\ref{fig:combined_failure_analysis}d. Across both benchmarks, these observations identify preservation of scientific inputs as a recurring difference between the evaluated systems. They do not establish which architectural component caused that difference, because prompting, provider routes, and output-handling mechanisms varied across configurations.

\begin{figure*}[ht]
    \centering
    \includegraphics[width=\textwidth]{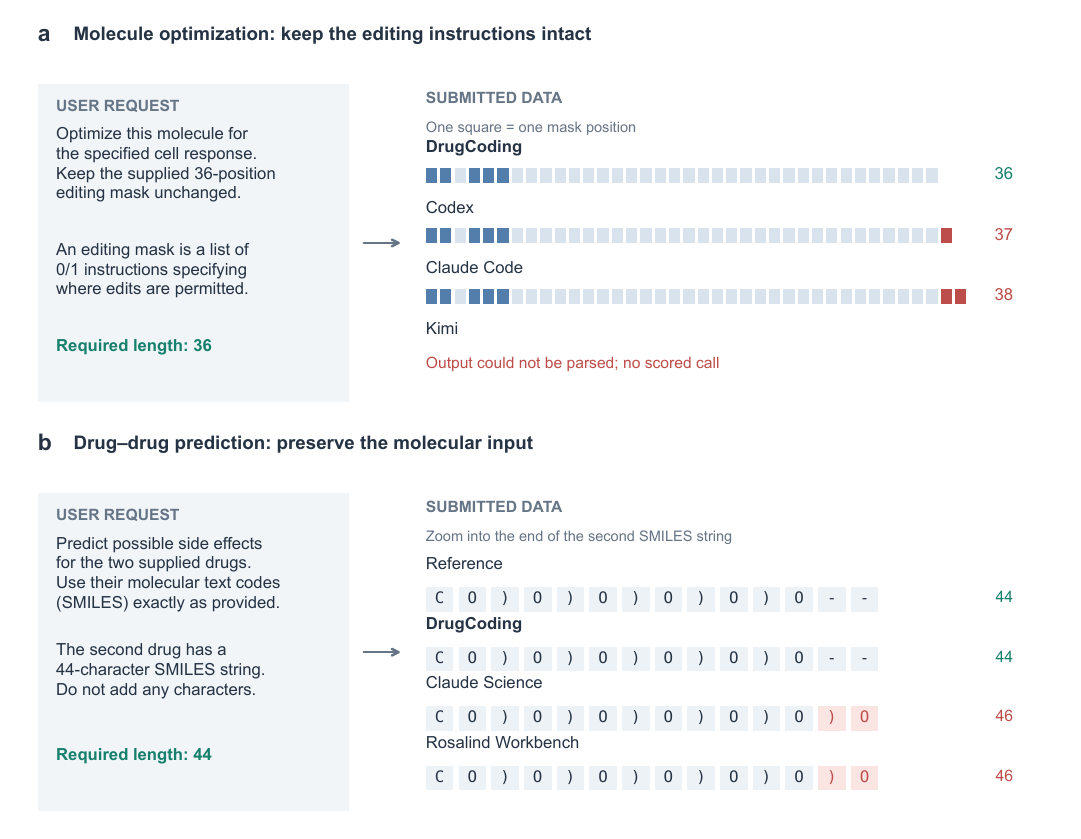}
    \caption{Illustrative cases of scientific input preservation. \textbf{a}, Molecule optimization with a supplied 36-element editing mask. Squares represent individual entries: blue denotes 1, pale blue denotes 0, and red marks added positions. \textbf{b}, Drug-drug interaction prediction with supplied SMILES strings. The aligned excerpt shows two extra characters appended by Claude Science and Rosalind Workbench; red highlights the additions, and dashes denote display alignment gaps.}
    \label{fig:combined_case_studies}
\end{figure*}

\subsection*{Preserving Scientific Inputs: Case Studies}
To illustrate how small changes to scientific inputs affect task correctness, we examined two cases involving an editing mask and a molecular string in Figure~\ref{fig:combined_case_studies}. Each case compares the submitted data against an identical benchmark reference across the displayed agents.

The molecule-optimization case required preserving a 36-element editing mask alongside the supplied molecule, cell identifier, and target response value, as shown in Figure~\ref{fig:combined_case_studies}a. DrugCoding reproduced the complete reference call, including every mask entry. Codex and Claude Code selected the correct operation but extended the mask to 37 and 38 elements, respectively. Kimi produced an output that could not be parsed into a scored call. This example shows that identifying the intended operation does not ensure correctness when its execution also depends on preserving structured constraints.
The drug-drug interaction case required passing two supplied SMILES strings without alteration in Figure~\ref{fig:combined_case_studies}b. DrugCoding preserved both strings exactly. Claude Science and Rosalind Workbench instead appended \texttt{)O} to the second string, increasing its length from 44 to 46 characters. The aligned excerpt highlights how two added characters caused an otherwise similar submission to differ from the reference. These cases highlight DrugCoding's ability to preserve scientific inputs and produce correctly structured submissions across the illustrated tasks.

\section*{Conclusion}
In this work, we presented DrugCoding, a scientific programming agent system that connects research objectives to computational workflows through explicit task dependencies, adaptive scientific-model routing, coordinated specialist agents, and traceable execution and validation. Across the evaluated configurations, DrugCoding achieved an average Pass@1 of 97.35\% on the full multi-turn benchmark and 97.92\% on the challenging subset, outperforming the compared coding and scientific agents on average. Failure analysis and case studies identified scientific input preservation and structured-output reliability as recurring strengths, while the ablation study showed that the benefits of multi-agent coordination varied across language models. These findings support DrugCoding's ability to translate scientific requests into correctly structured computational submissions and motivate further evaluation of how this reliability carries through to executed workflows and validated scientific outcomes.

\bibliographystyle{unsrturl}
\bibliography{reference}

\end{document}